\documentclass[runningheads]{llncs}
\usepackage{graphicx}

\usepackage{graphicx}
\usepackage{amsmath}
\usepackage{amssymb}
\usepackage{booktabs}
\usepackage{multirow}
\usepackage{float}
\usepackage{multicol}
\usepackage{tikz}
\usepackage{comment}
\usepackage{amsmath,amssymb} 
\usepackage{color}
\usepackage{subcaption}
\usepackage[misc]{ifsym}

\newcommand{\etal}{\textit{et al}.}

\usepackage[accsupp]{axessibility}  

\usepackage[pagebackref,breaklinks,colorlinks]{hyperref}
\usepackage[capitalize]{cleveref}
\crefname{section}{Sec.}{Secs.}
\Crefname{section}{Section}{Sections}
\Crefname{table}{Table}{Tables}
\crefname{table}{Tab.}{Tabs.}

\begin{document}
\pagestyle{headings}
\mainmatter
\def\ECCVSubNumber{1020}  

\title{Hierarchical Memory Learning for Fine-Grained Scene Graph Generation} 

\titlerunning{HML for SGG}
%

%
\authorrunning{Y. Deng et al.}
%

\author{Youming Deng$^{1}$\index{Deng, Youming} \quad Yansheng Li$^{1}$(\Letter)\index{Li, Yansheng} \quad Yongjun Zhang$^1$\index{Zhang, Yongjun} \quad Xiang Xiang$^2$\index{Xiang, Xiang}\\Jian Wang$^3$\index{Wang, Jian} \quad Jingdong Chen$^3$\index{Chen, Jingdong} \quad Jiayi Ma$^{4}$\index{Ma, Jiayi}}
\institute{$^1$School of Remote Sensing and Information Engineering, Wuhan University \\ $^2$School of Artificial Intelligence and Automation, Huazhong University of Science and Technology \quad $^3$Ant Group \\
$^4$ Electronic Information School, Wuhan University\\
}

\maketitle


\begin{abstract}
Regarding Scene Graph Generation (SGG), coarse and fine predicates mix in the dataset due to the crowd-sourced labeling, and the long-tail problem is also pronounced. Given this tricky situation, many existing SGG methods treat the predicates equally and learn the model under the supervision of mixed-granularity predicates in one stage, leading to relatively coarse predictions. In order to alleviate the impact of the suboptimum mixed-granularity annotation and long-tail effect problems, this paper proposes a novel Hierarchical Memory Learning (HML) framework to learn the model from simple to complex, which is similar to the human beings' hierarchical memory learning process. After the autonomous partition of coarse and fine predicates, the model is first trained on the coarse predicates and then learns the fine predicates. In order to realize this hierarchical learning pattern, this paper, for the first time, formulates the HML framework using the new Concept Reconstruction (CR) and Model Reconstruction (MR) constraints. It is worth noticing that the HML framework can be taken as one general optimization strategy to improve various SGG models, and significant improvement can be achieved on the SGG benchmark. 
\end{abstract}

\section{Introduction}

\begin{figure}[t]
    \centering

    \includegraphics[width=1\linewidth]{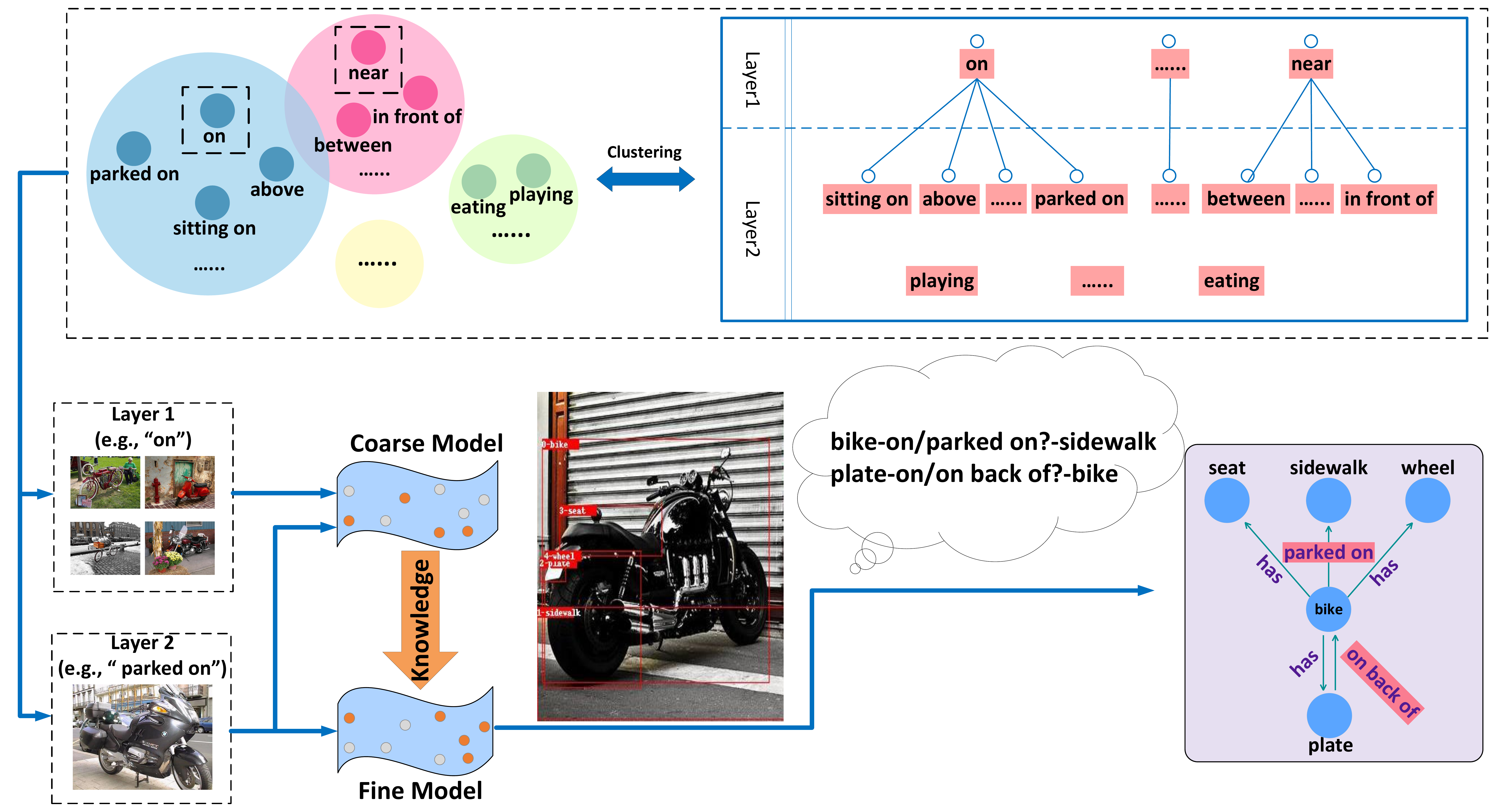}
    \caption{\textbf{Automatic Predicate Tree Construction and visualization of the HML Framework}. After constructing a hierarchical predicate tree via clustering, the model is trained with the HML framework to make the fine-grained predicate prediction.}
    \label{fig:pic1}
\end{figure}
The task of Scene Graph Generation (SGG)~\cite{c28} is a combination of visual object detection and relationship (i.e., predicate) recognition between visual objects. It builds up the bridge between computer vision and natural language. SGG receives increasing attention since an ideal informative scene graph has a huge potential for various downstream tasks such as image caption~\cite{c19,c29} and VQA~\cite{c18,c21}. To pursue the practical application value, SGG models keep working towards generating an informative scene graph where the fine-grained relationship between two objects should be predicted. Earlier works like~\cite{c3,c43,c4} only design models for feature refinement and better representation. However, they ignore the dataset and task properties, limiting the performance. In order to deal with the long-tail effect and various biases~\cite{c7} within the dataset, recent works including~\cite{c54,c53,c45,c7} move towards designing the learning framework to improve the overall performance of several classic SGG models. Even with this progress, making the fine-grained predicate prediction is still challenging. 

Generally speaking, two factors lead to a frustrating result. The first is the mixed-granularity predicates caused by artificial subjective annotation. The predicate recognition in the dataset is much trickier than the image classification tasks. For instance, although Microsoft coco~\cite{c48} has super categories, it does not require a model to have the ability to make a different prediction like ``bus''  and``vehicle'' for a visually identical object. However, in the SGG task, a more general predicate like ``on'' and a more informative one like ``parked on'' will be learned and predicted simultaneously. Under this training condition, it is a dilemma for models since machines cannot understand why almost identical visual features have different annotations and require them to make different prediction results. The second one is the long-tail effect which exists objectively in nature. Some of the dominant predicate classes are almost 1,000 times as many as less-frequent ones, leading to a bad performance on those less frequent predicates. Too many general predicates like ``on'' in training will lead to insufficient training for less-frequent ones like ``eating'' and drifting preference away from fine ones such as ``parked on''. Some methods, including re-weighting and re-sampling, seem to be suitable choices. However, due to the hierarchical annotation in the dataset, the improvement seems to be limited according to~\cite{c7}.

When training the deep network, problems like the long-tail effect and mixed-granularity annotation are universal but catastrophic to the deep network training. In contrast, humans seem capable of handling these complicated problems. As shown in cognitive psychology research~\cite{c47}, human beings appear to learn gradually and hierarchically. Coincidentally, the semantic structure of predicates in VG is similar to our real-life naming, consisting of mixed-granularity information. For instance, parents often teach their kids to recognize ``bird'' first and then go for the specific kinds like ``mockingbird'' or ``cuckoo''. Inspired by this, we realize that it is plausible to design a hierarchical training framework to resolve the abovementioned problems by imitating human learning behavior.

This work proposes a Hierarchical Memory Learning (HML) framework for hierarchical fashion training with the abovementioned consideration. At the very beginning, we cluster predicates, establish a hierarchical tree in \cref{fig:pic1} and separate the dataset by the tree layers without any extra manual annotation. To realize hierarchical training, Concept Reconstruction (CR) is used to inherit the previous model's predicate recognition ability by imitating its output. For a similar purpose, Model Reconstruction (MR) directly fits the parameters in the previous model as a stronger constraint. Under this training scenario, the model gets less chance to confront the previously discussed dilemma and is much easier to train with a relatively small and balanced fraction of predicates.

The proposed HML framework is a general training framework and can train any off-the-shelf SGG model. \cref{fig:pic1} shows the scene graph generated by the hierarchical training scenario. The scene graph predicted by the model trained with the HML framework is more comprehensive and fine-grained. The predicted relationships such as ``bike-parked on-sidewalk'' and ``plate-on back of-bike'' are more informative and meaningful than``bike-on-sidewalk'' and ``plate-on-bike''.

The main contributions of this work can be summarized as follows:
\begin{itemize}
\item Inspired by human learning behavior, we propose a novel HML framework, and its generality can be demonstrated by applying it to various classic models.
\item We present two new CR and MR constraints to consolidate knowledge from coarse to fine.
\item Our HML framework overperforms all existing optimization frameworks. Besides, one standard model trained under HML will also be competitive among various SGG methods with the trade-off between fine and coarse prediction.
\end{itemize}

\section{Related Work}
\label{sec:rela}
\subsection{Scene Graph Generation}
SGG~\cite{c91,c28,c4} has received increasing attention in the computer vision community because of it's potential in various down-stream visual tasks~\cite{c30,c20,c29}. Nevertheless, the prerequisite is the generation of fine-grained and informative scene graphs. Recent works consider SGG mainly from three perspectives.

\noindent\textbf{Model Design}. Initially, some works designed elaborate structures for better feature refinement.~\cite{c28} leveraged GRUs to pass messages between edges and nodes.~\cite{c78} explored that the feature of objects and predicates can be represented in low-dimensional space, which inspired works like~\cite{c80,c79,c81}. ~\cite{c4} chose BiLSTM for object and predicate context encoding.~\cite{c3} encoded hierarchical and parallel relationships between objects and carried out a scoring matrix to find the existence of relationships between objects. Unfortunately, the improvement is limited to elaborate model design alone.


\noindent\textbf{Framework Formulation}. Later works tried to design the optimization framework to improve the model performance further. Based on causal inference,~\cite{c7} used Total Direct Effect for unbiased SGG.~\cite{c45} proposed an energy-based constraint to learn predicates in small numbers.~\cite{c50} formulated the predicate tree structure and used tree-based class-balance loss for training.~\cite{c50} and our work both focus on the granularity of predicates and share some similarities.

\noindent\textbf{Dataset Property}. The long-tail effect was particularly pronounced in VG, making studying this problem very important.~\cite{c54} utilized dynamic frequency for the better training.~\cite{c34} proposed a novel class-balance sampling strategy to capture entities and predicates distributions.~\cite{c53} sought a semantic level balance of predicates.~\cite{c6} used bipartite GNN and bi-level data re-sampling strategy to alleviate the imbalance. However, another problem (mixed-granularity annotation) in the dataset is not fully explored, which inspires this work. Our concurrent work~\cite{c90} also borrowed the incremental idea to overcome this problem. We add semantic information for the separation and stronger distill constraint for better head knowledge preserving, while~\cite{c90} learns better at the tail part.

\subsection{Long-Tail Learning}
Only a few works like~\cite{c54,c34,c53,c6} cast importance on the long-tail effect in VG. In fact, many long-tail learning strategies can be used in SGG. The previous works tackling the long-tail effect can be roughly divided into three strategies. 

\noindent\textbf{Re-sampling}. Re-sampling is one of the most popular methods to resolve class imbalance. Simple methods like random over or under-sampling lead to overfitting the tail and degrading the head. Thus, recent work like~\cite{c60,c61,c62,c63} monitored optimization process of depending only instance balance.

\noindent\textbf{Cost-sensitive Learning}. Cost-sensitive learning realizes class balance by adjusting loss for different
classes during training.~\cite{c55,c48,c65} leveraged label frequency to adjust loss and prediction during the training.~\cite{c66} regarded one positive sample as a negative sample for other classes in calculating softmax or sigmoid cross-entropy loss. Other works~\cite{c68,c67} tried to handle the long-tail problem by adjusting distances between representation features and the model classifier for different classes.

\noindent\textbf{Transfer or Incremental Learning}. Those methods help to transfer information or knowledge from head to tail and enhance models' performances.~\cite{c72,c71} proposed a guiding feature in the head to augment tail learning.~\cite{c73} learned to map few-shot model parameters to many-shot ones for better training. Works like~\cite{c35,c74,c36} helped to distill knowledge directly from the head to the tail.

\section{Approach}
\label{sec:approach}
We will first introduce how to automatically construct a hierarchical predicate tree via clustering (\cref{sec:Construction of Hierarchical Tree}). And then turn back to explain our HML framework in (\cref{sec:HML}), along with loss formulation for CR (\cref{sec:CRL}) and MR (\cref{sec:MRL}).

\subsection{Soft Construction of Predicate Tree}
\label{sec:Construction of Hierarchical Tree}
In order to form a predicate tree for our training scenario, we firstly embed all 50 predicates in the dataset into feature vectors with the pre-trained word representation model in~\cite{c59}. 
After that, motivated by reporting bias~\cite{c8}, we cluster predicates into different groups and do some soft manual pruning (e.g., re-classifying some mistakes into different groups). We finally pick up the top-K frequent predicates within each group as the first K layers of the tree.

As for clustering, we choose the traditional distributed word embedding (DWE) algorithm~\cite{c59}, since we wish to eliminate the context from objects or subjects which can provide extra information~\cite{c4} to the predicate embedding.
We iterate through all 50 predicates from frequent to less-frequent for clustering. The first predicate is automatically divided into the first group and records its embedding vector to be the initial value of the first group representation $R_{1}=DWE(x_{1})$. As for the following ones, we calculate the cosine distance among all current group representations:
\begin{equation}
    SS^{ij}=\frac{R_{i}\cdot DWE(x_{j})}{||R_{i}||\times ||DWE(x_{j})||},
\end{equation}
where $SS^{ij}$ is the semantic similarity between current iterated predicate $x_{j}$ and $i^{th}$ group representations $R_{i}$. $DWE(x)$ represents distributed word embedding function on the predicate. Max cosine distance $SS_{max}^{ij}$ will be recorded and compared with the empirical threshold $T_{SS}$ whose setting is mentioned in \cref{sec:imple}. If $SS^{ij}$ is larger than the threshold, the currently iterated predicate $x_{j}$ will be added to the existing group. Otherwise, we create a new group for it. The group representations will be updated in the following rule:
\begin{equation}
\label{eq:thre}
\left\{
    \begin{array}{l}
    R_{N+1}=DWE(x_{j}),  SS_{max}^{ij} < T_{SS}\\
    R_{i}=\frac{n\cdot R_{i} + DWE(x_{j})}{n + 1},  SS_{max}^{ij} \geq T_{SS}\\
    \end{array}
\right.,
\end{equation}
where $N$ is the current number of groups and $n$ is the number of predicates in the $i^{th}$ group.

After clustering, the most frequent predicates are assigned to the first layer, the second frequent predicates are assigned to the second layer, etc. It is worth noticing that during this clustering, some human actions such as ``looking at", ``playing", ``says", ``eating", and ``walking in" will become a single group as one single predicate. We automatically divide them into the last layer since those human action predicates are almost 100 to 1000 times less than predicates in the other layer. The most suitable number of layers depends on the dataset itself. \cref{sec:ablation} analyzes the best layer number for the VG dataset.

\subsection{Hierarchical Memory Learning Framework}
\label{sec:HML}

Most SGG models comprise two steps. In the beginning, an image is fed into an ordinary object detector to get bounding boxes, corresponding features of these regions, and the logits over each object class. These detection results are used to predict the scene graph in the next step. The feature of a node is initialized by box features, object labels, and position. Some structures like LSTM~\cite{c1,c4} are used to refine nodes' features by passing and incorporating the messages. After that, the object labels are obtained directly by refined feature, while the predicates are predicted from the union features refined by the structures of BiLSTM~\cite{c4}, BiTreeLSTM~\cite{c1}, GRU~\cite{c5}, or GNN~\cite{c6}. 

\begin{figure}[t]
  \centering
  \includegraphics[width=1\linewidth]{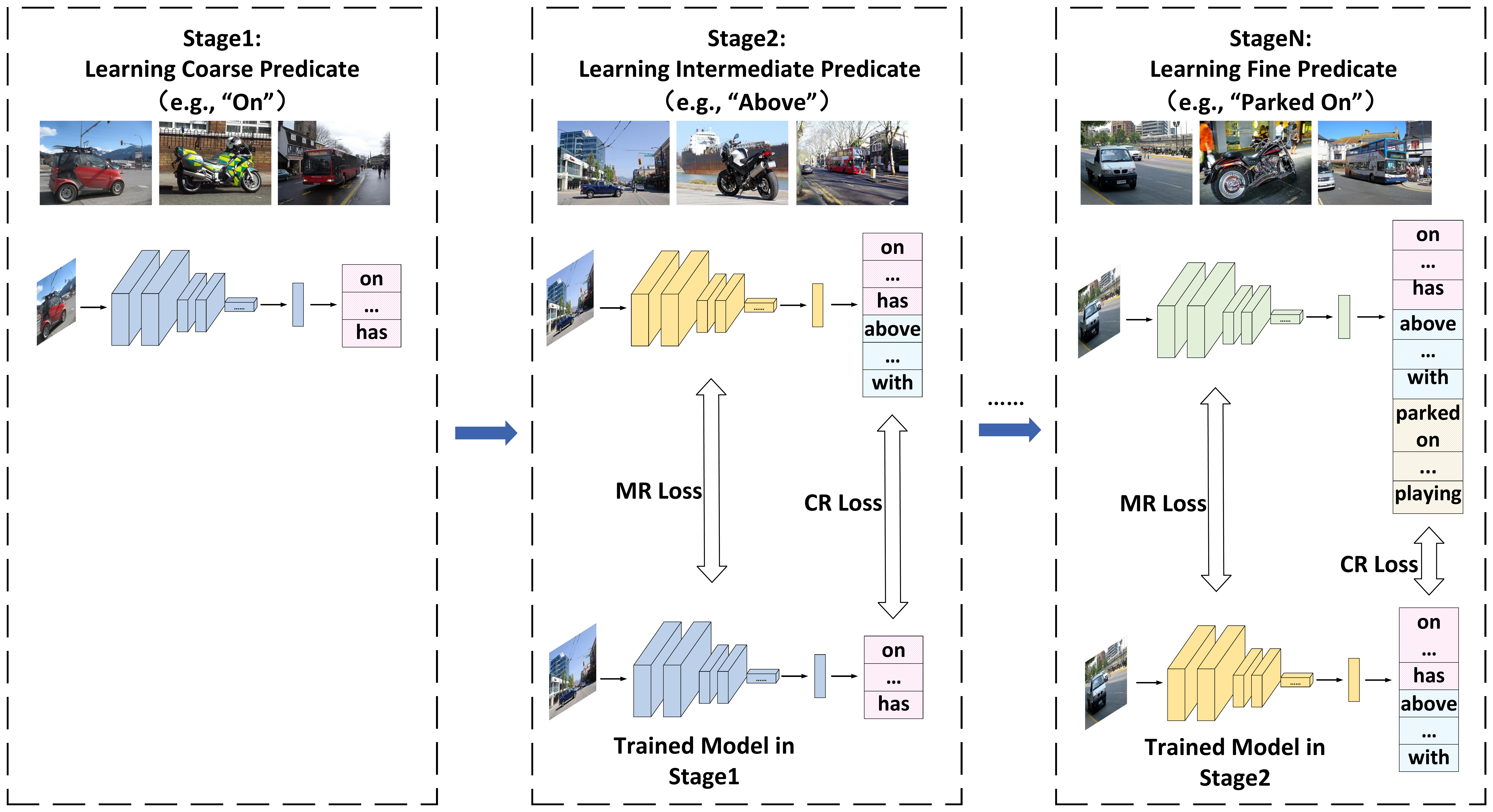}
  
  \caption{\textbf{Overview of HML Framework}. We train the model in a coarse to fine fashion. In each step, the model calculates CR and MR for knowledge assimilation. Meanwhile, the importance scores and empirical Fisher Information Matrix are calculated after updating the model. The importance score and empirical Fisher Information Matrix are passed down at the end of each stage. For the VG dataset, we set the stage number to be 2 and explain in \cref{sec:ablation}.}
  \label{fig:pic5}
\end{figure}

Nevertheless, most models are still trained on the whole dataset at one time, making the task challenging. To address the long-tail effect and mixed-granularity annotation, we believe it is a better solution to disentangle semantic-confusing predicate classes by dealing with general relationships (e.g., ``on", ``has") and informative ones (e.g., ``sitting on", ``using") separately in different stages. In this training scenario, the model in its stage can only focus on a small fraction of predicates with relatively similar granularity and then congregate the knowledge from previous stages step by step.

Since the SGG task is similar to the human learning pattern, absorbing knowledge from coarse to fine, it is natural to model how humans learn. Given the model trained in the previous stage, the current training model needs to gain the ability to do well in previous classes while learning how to deal with new classes. A naive way is to sample some images in the last stage for review. Nevertheless, this strategy is unsuitable for our situation since it reintroduces mixed-granularity predicates. Thus, for better knowledge consolidation, we adopt \textbf{Concept Reconstruction} (\textbf{CR}) and \textbf{Model Reconstruction} (\textbf{MR}) which will be further explained in \cref{sec:CRL} and \cref{sec:MRL}. CR will be adopted to decrease the distance between the prediction logits produced by the two models. This process is similar to how human students imitate teachers to solve problems. Human brain cortical areas have different functional networks~\cite{c47}. It is the same for the parameters in an SGG model. MR respects the hypothesis that different parameters in a model serve for different relationship recognition. As is shown in \cref{fig:pic5}, in the second to $N^{th}$ stage, the model will fit the prediction and parameters for knowledge passing. At the same time, the gradient and parameters' change will be stored for the online update of the empirical Fisher Information Matrix (\cref{sec:MRL}) and importance scores (\cref{sec:MRL}).

Does a deep network have parameter redundancy? Some earlier work did answer this question. Hinton \etal~\cite{c9} came up with knowledge distillation for the first time to compress and transfer knowledge from a complicated model to a more compact one that is easier to deploy. Moreover, it was verified in~\cite{c10} that a cumbersome model can be compressed into a compact one. We assume that the whole model comprises many parameters based on these works. These parameters or ``activated'' neurons in an SGG network should target specific predicate classes, as shown in \cref{fig:pic1}. To verify this assumption, we compare the mean of importance scores, which will be further illustrated in \cref{sec:MRL} and verified in supplementary material from the experimental perspective, of all the parameters in each layer and find out that the values vary between different stages. This mechanism is similar to how the human brain works. Each region in the brain has its' own function and works together to finish complex tasks~\cite{c47}. After learning through all stages, the model will classify all classes with fine-grained preference.

The total loss for the HML framework is given by:
\begin{equation}
\label{eq:overall loss}
    \ell=\ell_{new}+ \ell_{CR}+\lambda \ell_{MR}, 
\end{equation}
where $\lambda$ is a hyper-parameter. $\ell_{CR}$ is Concept Reconstruction loss, and $\ell_{MR}$ is Model Reconstruction loss.
    
$\ell_{new}$ in \cref{eq:overall loss} is Class-Balance loss~\cite{c55} and used to learn current stage predicates:
\begin{equation}
    \ell_{new}=-W_B\sum_{i=1}^{C}{y_i\log{\frac{e^{z_i}}{\sum_{j=1}^{C}e^{z_j}}}},
\end{equation}where $z$ represents the model's output, $y$ is the one-hot ground-truth label vector, and $C$ is the number of predicate classes. $W_B=\frac{1-\gamma}{1-\gamma^{n_i}}$, $\gamma$ denotes the hyper-parameter that represents the sample domain, and $n_i$ denotes number of predicate $i$ in current stage.

\subsection{Concept Reconstruction Loss}
\label{sec:CRL}
In order to prevent activation drift~\cite{c10}, CR is applied. It is an implicit way to keep the prediction of previous predicates from drifting too much. Thus, we need to find the distance between two predictions from different stages of the same visual relationship and reduce it. The CR loss can be represented by:
\begin{equation}
    \ell_{CR}(X_{n}, Z_{n})=\frac{\sum_{i=1}^{N_n}\sum_{j=1}^{C_{old}}\left(\text{Softmax}\left(x_n^{ij}\right)-\text{Softmax}\left(z_n^{ij}\right)\right)^2}{N_n},
\end{equation}where $x_n^i$ and $z_n^i$ are the output logits vector for the prediction. $N_n$ is the number of outputs. We choose L2 distance~\cite{c51} as the distance metric. Compared with the traditional loss function such as L1 loss and cross-entropy loss with the soft label, L2 loss is a stronger constraint but is sensitive to outliers. Fortunately, since the training process is coarse to fine, the representations will not drastically deviate, making L2 loss practical. With the consideration mentioned earlier, L2 distance is used in CR, receiving better performance in experiments. This is also verified in~\cite{c16}. More ablation results of CR can be found in the supplementary material.

In a word, CR is used to help the current model learn how to make the same prediction as the previous model.

\subsection{Model Reconstruction Loss}
\label{sec:MRL}
The parameters of the model determine the ability to recognize visual relationships. Thus, it is a more straightforward way to learn directly from parameters. A feasible solution is to determine which parameters are crucial in the previous stage classification and fit them with greater attention in the following stage. 

KL-divergence is a mathematical statistics measure of how a probability distribution is different from another one~\cite{c11}. KL-divergence, denoted as in the form of $D_{KL}(p_\theta||p_{\theta+\Delta\theta})$, can also be used to calculate the difference of the conditional likelihood between a model at $\theta$ and $\theta+\Delta\theta$. Since changes of parameters are subtle (i.e., $\Delta\theta\rightarrow0$) during the optimization, we will get the second-order of Taylor approximation of KL-divergence, which is also the distance in Riemannian manifold induced by Fisher Information Matrix~\cite{c52} and can be written as $D_{KL}(p_\theta||p_{\theta+\Delta\theta})\approx\frac{1}{2}\Delta\theta^\top F_\theta\Delta\theta$, where the $F_\theta$ is known as empirical Fisher Information Matrix~\cite{c12} at $\theta$ and the approximate will be proved in supplementary material, is defined as:
\begin{equation}
    F_\theta=\mathbb{E}_{\left(\mathbf{x}, \mathbf{y}\right)\sim D}\left[\left(\frac{\partial \log p_\theta\left(\mathbf{y}\middle|\mathbf{x}\right)}{\partial\theta}\right)\left(\frac{\partial \log p_\theta\left(\mathbf{y}\middle|\mathbf{x}\right)}{\partial\theta}\right)^\top\right],
\end{equation}where $D$ is the dataset and $p_\theta\left(y\middle| x\right)$ is the log-likelihood. However, in practice, if a model has $P$ parameters, it means $F_\theta\in R^{P\times P}$, and it is computationally expensive. To solve this, we compromise and assume parameters are all independent, making $F_\theta$ diagonal. Then the approximation of KL-divergence looks like:
\begin{equation}
  D_{KL}(p_\theta||p_{\theta+\Delta\theta})\approx\frac{1}{2}\sum_{i=1}^{P}{F_{\theta_i}\Delta{\theta_i^2} },  
\end{equation}
where $\theta_i$ is the $i^{th}$ parameters of the model and $P$ is the total number of it.

$F_\theta$ will be updated in each iteration, following the rule:
\begin{equation}
    F_\theta^t = \frac{F_\theta^t + (t-1)\cdot F_\theta^{t-1} }{t},
\end{equation}
where $t$ is the number of iterations.

Although the empirical Fisher Information Matrix captures static information of the model, it fails to capture the influence of each parameter during optimization in each stage. Thus, we adopt the method in~\cite{c13} to search for essential parameters. Intuitively, if the value of a parameter changes a little in a single step, but it contributes a lot to the decrease of the loss, we think it is essential, at least for the current task. So, the importance of a parameter during an interval (from $t$ to $\Delta t$) can be represented as:
\begin{equation}
    \Omega_{raw}\left(\theta_i\right)=\sum_{t}^{t+\Delta t}\frac{{\Delta}\ell_{t}^{t+1}\left(\theta_i\right)}{\frac{1}{2}F_{\theta_i}^t{\left(\theta_i\left(t+1\right)-\theta_i\left(t\right)\right)^2+\epsilon}},
\end{equation}
\begin{equation}
    \Omega_t^{t+\Delta t}\left(\theta_i\right)=\sigma\left(\log_{10}{\frac{{{P\times\Omega}_{raw}\left(\theta_i\right)}}{\sum_{i=1}^{P}{\Omega_{raw}\left(\theta_i\right)}}}\right),
\end{equation}
where $\sigma$ is the sigmoid function, numerator ${\Delta}\ell_{t}^{t+1}\left(\theta_i\right)$ is the change of loss caused by $\theta_i$ in one step, $\epsilon>0$ aims to avoid the change of loss $\theta_i\left(t+1\right)-\theta_i\left(t\right)=0$, and the denominator is the KL-divergence of $\theta_i$ between $t$ and $t+1$. 

To be more specific, ${\Delta}\ell_{t}^{t+1}\left(\theta_i\right)$ represents how much contribution does $\theta_i$ make to decrease the loss. Since the optimization trajectory is hard to track, to find the change in loss caused by $\theta_i$, we need to figure out a way to split the overall loss form into the sum of each parameter's contributions. The solution is a first-order Taylor approximation:
\begin{equation}
\begin{aligned}
    \ell\left(\theta\left(t+1\right)\right)-\ell\left(\theta\left(t\right)\right)\approx-\sum_{i=1}^{P}\sum_{t=t}^{t+1}{\frac{\partial \ell}{\partial\theta_i}\left(\theta_i\left(t+1\right)-\theta_i\left(t\right)\right)}=-\sum_{i=1}^{P}{\mathrm{\Delta}\ell_{t}^{t+1}\left(\theta_i\right)},
\end{aligned}
\end{equation}where $\frac{\partial \ell}{\partial\theta_i}$ is the gradient of $\theta_i$ and $\theta_i\left(t+1\right)-\theta_i\left(t\right)$ is the value change of $\theta_i$ during a single step. If $\ell\left(\theta\left(t+1\right)\right)-\ell\left(\theta\left(t\right)\right) > 0 $, we set ${\Delta}\ell_{t}^{t+1}\left(\theta_i\right)$ to be $0$, since we consider only when the loss become smaller, a step of optimization can be regarded as effective.

The empirical Fisher Information Matrix is used twice. The first is to calculate the difference in probability distributions of two models in different stages, and the second is to find the changes of a model in a nearby iteration within a single stage.

So, after figuring out how important each parameter is, the MR loss can be written as:
\begin{equation}
    \ell_{MR}=\frac{\sum_{i=1}^{P}{\left({F_{\theta_i}^{k-1}+(\Omega}_{t_0}^{t_{0}+\Delta t})^{k-1}\left(\theta_i\right)\right)\left(\theta_i^{k}-\theta_i^{k-1}\right)^2}}{P},
\end{equation}where $P$ is the number of parameters for relationship prediction in the model and $k$ represents the current stage. $F_{\theta_i}^{k-1}$ and $(\Omega_{t_0}^{t_{0}+\Delta t})^{k-1}$ are both calculated in previous stage.

Fisher Information Matrix $F_{\theta}$ and importance scores $
\Omega_{t_0}^{t_{k-1}}$ are used to represent the importance of parameters from static and dynamic perspectives, respectively.

\section{Experiment}

\subsection{Dataset and Model}
\noindent\textbf{Dataset}. In the SGG task, we choose Visual Genome (VG)~\cite{c26} as the dataset for both training and evaluation. It comprises 75k object categories and 40k predicate categories. However, due to the scarcity of over 90$\%$ predicates are less than ten instances, we applied the widely accepted split in~\cite{c17,c1,c4}, using the 150 highest frequency objects categories and 50 predicate categories. The training set is set to be 70$\%$, and the testing set is 30$\%$, with 5k images from the training set for finetuning.~\cite{c7}. 

\noindent\textbf{Model}. We evaluate HML framework on three models and follow the setting in~\cite{c89}: MOTIFS~\cite{c4}, Transformer~\cite{c25,c22}, and VCTree~\cite{c1}.

\subsection{Evaluation}
\noindent\textbf{Sub-Tasks}: (1) \textbf{Predicate Classification}: given images, object bounding boxes, and object labels, predicting the relationship labels between objects. (2) \textbf{Scene Graph Classification}: given images and object bounding boxes, predicting object labels and relationship labels between objects. (3) \textbf{Scene Graph Detection}: localizing objects, recognizing objects, and predicting their relationships directly from images.


\begin{table}[t]

    \begin{subtable}[h]{1\textwidth}
        \centering
        
        \scalebox{0.7}{

            \begin{tabular}{ccccccccccc}
    \toprule
    \multicolumn{1}{c}{\multirow{2}[1]{*}{Model}} &
    \multicolumn{1}{c}{\multirow{2}[1]{*}{Framework}} &\multicolumn{3}{c}{Predicate Classification} & \multicolumn{3}{c}{   Scene Graph Classification} & \multicolumn{3}{c}{   Scene Graph Detection} \\
\cmidrule{3-11}    \multicolumn{2}{c}{} & \multicolumn{1}{c}{mR@20} & \multicolumn{1}{c}{mR@50} & \multicolumn{1}{c}{mR@100} & \multicolumn{1}{c}{mR@20} & \multicolumn{1}{c}{mR@50} & \multicolumn{1}{c}{mR@100} & \multicolumn{1}{c}{mR@20} & \multicolumn{1}{c}{mR@50} & \multicolumn{1}{c}{mR@100} \\
    \midrule
    \multirow{4}[0]{*}{Transformer~\cite{c22}} 
    &Baseline & 14.1 & 17.9 & 19.4 & 8.2  & 10.1 & 10.8 & 6.3  & 8.5   & 10.1 \\
    &CogTree~\cite{c50} & 22.9  & 28.4  & 31.0    & 13.0    & 15.7  & 16.7  & 7.9   & 11.1  & 12.7 \\
    &BPL+SA~\cite{c53} &   26.7   &   31.9   &  34.2    &   \textbf{15.7}   &  18.5    &   19.4   &  \textbf{11.4}    &  14.8    & 17.1\\
    &HML(Ours) & \textbf{27.4} & \textbf{33.3} & \textbf{35.9} & \textbf{15.7} & \textbf{19.1} & \textbf{20.4} & \textbf{11.4} & \textbf{15.0} & \textbf{17.7}  \\
    \midrule
    \multirow{8}[0]{*}{MOTIFS~\cite{c4}} 
    &Baseline & 12.5 & 15.9 & 17.2 & 7.4  & 9.1  & 9.7  & 5.3  & 7.3  & 8.6 \\
    &EBM~\cite{c45}   & 14.2 & 18.0 & 19.5 & 8.2  & 10.2 & 11.0 & 5.7  & 7.7  & 9.3 \\
    &SG~\cite{c76} &14.5& 18.5& 20.2 &8.9& 11.2& 12.1& 6.4& 8.3& 9.2\\
    &TDE~\cite{c7}   & 18.5  & 25.5  & 29.1  & 9.8   & 13.1  & 14.9  & 5.8   & 8.2   & 9.8 \\
    &CogTree~\cite{c50} & 20.9  & 26.4  & 29.0    & 12.1  & 14.9  & 16.1  & 7.9   & 10.4  & 11.8 \\
    &DLFE~\cite{c54}  & 22.1  & 26.9  & 28.8  & 12.8  & 15.2  & 15.9  & 8.6   & 11.7  & 13.8 \\
    &BPL+SA~\cite{c53} &24.8& 29.7 &31.7 &14.0 &16.5 &17.5& 10.7 &13.5 &15.6\\
    &GCL~\cite{c90}&\textbf{30.5} &36.1 &38.2 &\textbf{18.0} &\textbf{20.8} &21.8 &\textbf{12.9} &\textbf{16.8} &\textbf{19.3}\\
    &HML(Ours) & 30.1 & \textbf{36.3} & \textbf{38.7} & 17.1 & \textbf{20.8} & \textbf{22.1} & 10.8 & 14.6 & 17.3 \\

    \midrule
    \multirow{8}[0]{*}{VCTree~\cite{c3}}
    &Baseline & 13.4 & 16.8 & 18.1 & 8.5  & 10.5 & 11.2 & 5.9   & 8.2  & 9.6 \\
    &EBM~\cite{c45}   & 14.2  & 18.2 & 19.7 & 10.4  & 12.5 & 13.5 & 5.7  & 7.7  & 9.1 \\
    &SG~\cite{c76} &   15.0   &   19.2   &   21.1   &   9.3   &   11.6   &   12.3   &  6.3    &   8.1   & 9.0\\
    &TDE~\cite{c7}   & 18.4  & 25.4  & 28.7  & 8.9   & 12.2  & 14.0    & 6.9   & 9.3   & 11.1 \\
    &CogTree~\cite{c50} & 22.0    & 27.6  & 29.7  & 15.4  & 18.8  & 19.9  & 7.8   & 10.4  & 12.1 \\
    &DLFE~\cite{c54}  & 20.8  & 25.3  & 27.1  & 15.8 & 18.9 & 20.0 & 8.6   & 11.8  & 13.8 \\
    &BPL+SA~\cite{c53} &   26.2   &   30.6   &   32.6   &   17.2   &   20.1   &   21.2   &   10.6   &   13.5   & 15.7\\
    &GCL~\cite{c90}&\textbf{31.4} &\textbf{37.1} &39.1 &19.5 &22.5 &23.5 &\textbf{11.9} &\textbf{15.2}&\textbf{17.5}\\
    &HML(Ours) & 31.0 & 36.9 & \textbf{39.2} & \textbf{20.5} & \textbf{25.0} & \textbf{26.8} & 10.1 & 13.7 & 16.3 \\
    \bottomrule
    \end{tabular}%
    
    }
    
        \caption{\textbf{Comparison between HML and various optimization frameworks}.}
        \label{tab:quality up}
    \end{subtable}
   
    \begin{subtable}[h]{1\textwidth}
        \centering
        
        \scalebox{0.7}{
            \begin{tabular}{cccccccc}
        \toprule
        \multirow{2}[2]{*}{Model+Framework} & \multicolumn{2}{c}{Predicate Classification} & \multicolumn{2}{c}{Scene Graph Classification} & \multicolumn{2}{c}{Scene Graph Detection} &
        \multirow{2}[2]{*}{Mean@50/100}\\
    \cmidrule{2-7}          & mR@50/100 & R@50/100 & mR@50/100 & R@50/100 & mR@50/100 & R@50/100 \\
        \midrule
        MOTIFS-TDE~\cite{c7}      &   25.5/29.1     &  46.2/51.4    &   13.1/14.9  &  27.7/29.9  &   8.2/9.8   & 16.9/20.3  & 22.9/25.9\\
        MOTIFS-DLFE~\cite{c54}&26.9/28.8&\textbf{52.5}/\textbf{54.2} &15.2/15.9&\textbf{32.3}/\textbf{33.1} &11.7/13.8&\textbf{25.4}/\textbf{29.4}&27.3/29.2\\
        Transformer-CogTree~\cite{c50}     &  28.4/31.0&  38.4/39.7&  15.7/16.7&  22.9/23.4&  11.1/12.7&  19.5/21.7&22.7/24.2\\
        PCPL~\cite{c75}  &  35.2/37.8&50.8/52.6& 18.6/19.6 &27.6/28.4& 9.5/11.7 &14.6/18.6&26.1/28.1\\
        DT2-ACBS~\cite{c34}     & 35.9/39.7&23.3/25.6 &24.8/\textbf{27.5}& 16.2/17.6 &\textbf{22.0}/\textbf{24.4} &15.0/16.3&22.9/25.2\\
        SHA-GCL~\cite{c90}     & \textbf{41.6}/\textbf{44.1}&35.1/37.2 &23.0/24.3& 22.8/23.9 &17.9/20.9 &14.9/18.2&25.9/28.1\\
        \midrule
        Transformer-HML(Ours) &   33.3/35.9    &   45.6/47.8    &    19.1/20.4   &   22.5/23.8    &     15.0/17.7  &  15.4/18.6 & 25.2/27.4\\
        MOTIFS-HML(Ours) &   
        36.3/38.7    &   47.1/49.1   &   20.8/22.1    &  26.1/27.4   &    14.6/17.3    &  17.6/21.1 & 27.1/29.3\\
        VCTree-HML(Ours) &   36.9/39.2    &   47.0/48.8    &    \textbf{25.0}/26.8   &  27.0/28.4     &   13.7/16.3    & 17.6/21.0 &\textbf{27.9}/\textbf{30.1}\\
        \bottomrule
        \end{tabular}%
        }
        \caption{\textbf{A more comprehensive comparison between HML and various SOTAs}.}
        \label{tab:quality down}
    \end{subtable}
    \caption{\textbf{Result of Relationship Retrieval mR@K~\cite{c3} and R@K}.}
    \label{tab:quali result}%
\end{table}%

\noindent\textbf{Relationship Recall}. We choose \textbf{Mean Recall@K (mR@K)}~\cite{c3} as a metric to evaluate the performance of SGG models. As is shown in~\cite{c7}, regular Recall@K (R@K) will lead to the reporting bias due to the imbalance that lies in the data (e.g., a model that only correctly classifies the top 5 frequent predicates can reach 75\% of Recall@100). Thus, we introduce \textbf{Mean@K} which calculates the average score of all three sub-tasks R@K and mR@K under identical K. Mean@K is a metric to evaluate overall performance on both R@K and mR@K. We will further explain the reason and necessity of using this metric in the supplementary material.

\subsection{Implementation Details}
\label{sec:imple}


\noindent\textbf{Object Detector}. We pre-train a Faster-RCNN with  ResNeXt-101-FPN~\cite{c23} and freeze the previously trained weight during the SGG training period. The final detector reached 28 mAP on the VG test set.

\noindent\textbf{Relationship Predictor}. The backbone of baseline models is replaced with an identical one, and hierarchical trained in the same setting. We set the batch size to 12 and used SGD optimizer with an initial learning rate of 0.001, which will be decayed by 10 after the validation performance plateaus. The experiment was carried out on NVIDIA TITAN RTX GPUs.

\noindent\textbf{Hierarchical Predicate Tree Construction}. For Visual Genome (VG), the whole predicates are split into two disjoint subsets (tasks) following the construction in \cref{sec:Construction of Hierarchical Tree}. The threshold is set to be $T_{SS} \in [0.65, 0.75]$ in \cref{eq:thre}. However, due to the small semantic variance within VG, this Predicate Tree Construction degenerates to simply separating frequent ``on" or ``has" with minority ones, similar to the separation in~\cite{c90}. We believe a larger semantic variance dataset will need this kind of semantic information for more reasonable separation.

\noindent\textbf{Hierarchical Training}. In our experiment, two identical models will be trained separately in two stages. Moreover, the first model will be initialized randomly, and the previous stage models will not be used to initialize the following model. If not, the model will meet the intransigence~\cite{c37} problem. Besides, we set the max iterations to 8000 and 16000 in the first and second stages. Nevertheless, models will usually converge around 8000 in both stages. The $\lambda$ in an overall loss is set to 0.5 for the best performance, as is shown in \cref{tab:ab on ln}.


\begin{figure}[t]
    \centering
    \includegraphics[width=1\linewidth]{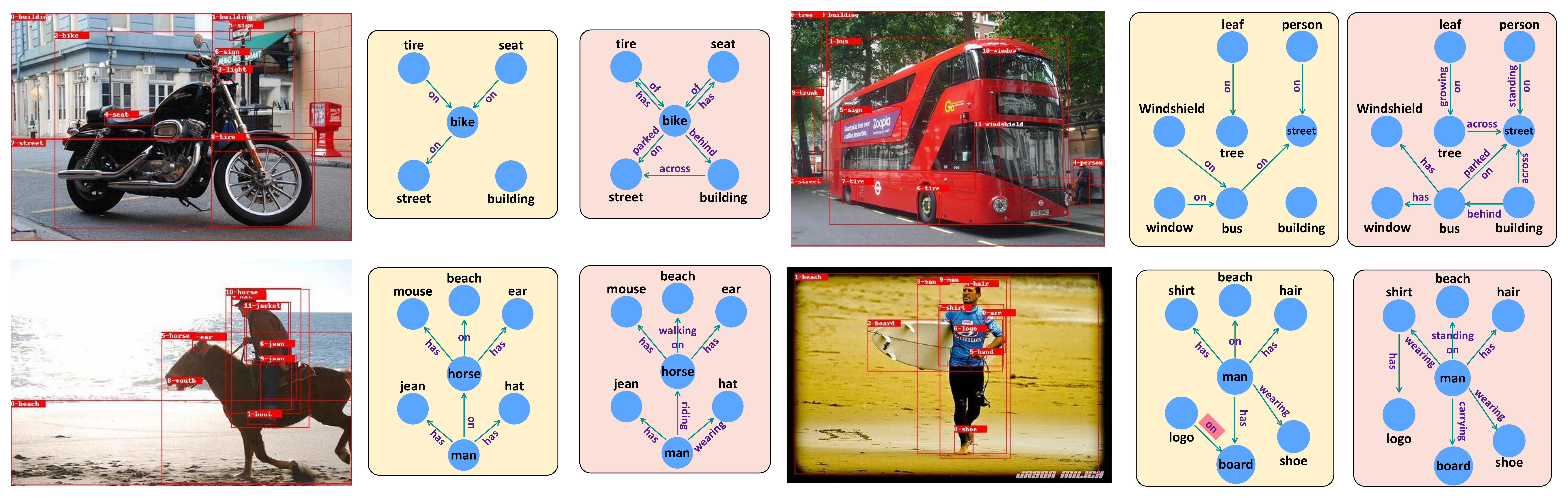}
    
    \caption{\textbf{Qualitative Results}. The basic model generates yellow scene graphs, and the same basic model predicts pink ones under HML training.}
    \label{fig:pic6}
\end{figure}

\subsection{Quantitative Study}
In \cref{tab:quality up}, our HML framework can consistently outperform the existing frameworks under various fundamental models. By applying the HML framework to different models, we notice that models will drastically improve the Mean Recall for all three sub-tasks. The improvement scale is similar to different models such as Motif, Transformer, and VCTree. In \cref{tab:quality down}, we compare HML with various SOTAs, which report R@K and mR@K simultaneously. After calculating Mean@K, it turns out that HML can make the fine-grained prediction (demonstrated by mR@K) and keep clear boundaries of general predicates (demonstrated by R@K).

In \cref{tab:up}, we notice a vital decrease in R@K. After analyzing each predicate, we figure out that the decline mainly comes from the drop of general predicates such as ``on'' and ``has''. After HML training, the mR@100 of ``on'' and ``has'' dropped from 79.98 to 31.01 and 81.22 to 58.34. The model's preference for fine-grained prediction causes this decrease. Thus, we replace the fine-grained one with a general one and recalculate R@100, and the results of ``on'' and ``has'' re-bounce to be 71.72 and 86.91, respectively. Works like~\cite{c54,c34,c7,c75,c50} compared in \cref{tab:quality down} can also make relatively fine-grained prediction (i.e., relatively high mR@K) but all suffer from decrease of R@K partially due to the reason mentioned above.

\begin{table*}[t]
    \centering

    \scalebox{1}{
    \begin{tabular}{ccccccccc}
  \toprule
  \multirow{2}[4]{*}{Model} & \multirow{2}[4]{*}{CR} & \multirow{2}[4]{*}{MR} & \multicolumn{6}{c}{Predicate Classification} \\
\cmidrule{4-9}          &       &       & mR@20 & mR@50 & mR@100& R@20  & R@50  & R@100  \\
  \midrule
    \multirow{4}[2]{*}{Transformer~\cite{c22}} 
        &       &       & 14.13 & 17.87  & 19.38 & 58.79 & 65.29& 67.09 \\
        & $\checkmark$ &      & 24.14 & 29.27  & 31.22  & 29.1& 36.16& 38.57   \\
        &       & $\checkmark$ & 23.32 & 29.34 & 32.20 & 38.80  & 46.48 & 48.87 \\
        & $\checkmark$ & $\checkmark$ & \textbf{27.35} & \textbf{33.25} & \textbf{35.85} & 38.81 & 45.61 & 47.78 \\
  \midrule
  \multicolumn{1}{c}{\multirow{4}[2]{*}{MOTIFS~\cite{c4}}} &             &        & 12.54 & 15.89  & 17.19 & 59.12 & 65.45 & 67.20\\
        & $\checkmark$ &        & 24.69& 30.00    & 32.79 & 33.92 & 41.34  & 43.95\\
        &       & $\checkmark$ & 21.56 & 27.43  & 30.05& 44.30  & 51.87 & 54.14 \\
        & $\checkmark$ & $\checkmark$ & \textbf{30.10}  & \textbf{36.28}& \textbf{38.67}& 40.52 & 47.11 & 49.08  \\
  \midrule

  \multirow{4}[2]{*}{VCTree~\cite{c3}} 
        &       &       & 13.36 & 16.81 & 18.08 & 59.76 & 65.48 & 67.49 \\
        & $\checkmark$ &       & 22.32  & 29.34& 32.20  & 27.86& 35.21 & 37.73 \\
        &       & $\checkmark$  & 22.84   & 28.94  & 31.48& 43.81& 51.40& 53.74 \\
        & $\checkmark$ & $\checkmark$ & \textbf{31.04} & \textbf{36.90} & \textbf{39.21} & 40.28 & 46.47 & 48.36\\
  
  \bottomrule
  \end{tabular}%
}
    
    \caption{\textbf{Ablation on CR and MR}. We explore the functionality of CR loss and MR loss.}
    \label{tab:up}%
    
  \end{table*}%
  
\subsection{Qualitative Study}
We visualize the qualitative result generated by original MOTIFS and MOTIFS trained with the HML framework in \cref{fig:pic6}. Compared with the original model, the model trained under HML will make informative predictions such as ``parked on'', ``growing on'', ``standing on'', ``riding'', and ``walking on'' instead of the general one ``on''. Also, our model will tend to make fine-grained predictions such as ``wearing'' and ``carrying'' instead of ``has''. Besides, since we train a model hierarchically, the model will obtain the ability to capture tail part predicates such as ``building-across-street'' and position predicates such as ``building-behind-bus'' and ``tree-behind-bus''. Qualitative results with three stages are shown in the supplementary material.

\begin{table*}[t]
  
    \begin{subtable}[h]{0.58\textwidth}
    
      \centering
      
      \scalebox{1.14}{
        \begin{tabular}{ccccc}
        
    \toprule
    \multirow{2}[2]{*}{layer} & \multicolumn{3}{c}{Predicates Classification} &  
    \multirow{2}[2]{*}{time (hr)}\\
    \cmidrule{2-4}          & mR@20 & mR@50 & mR@100 \\
    \midrule
        1  &     12.54 & 15.89 & 17.19& 17.85\\
        2  & \textbf{30.10} & \textbf{36.28} & \textbf{38.67}& 29.43\\
        3  &     25.24 & 31.95 & 34.44& 44.65\\
        4  &     15.66  &    21.96   & 25.32 & 60.13\\
    \bottomrule
    
    \end{tabular}%
      }
    
    \caption{\textbf{Ablation of Layer Number}}
    \label{tab:ab on ln}
    \end{subtable}
    \begin{subtable}[h]{0.39\textwidth}
      \centering
      
      \scalebox{1}{
        \begin{tabular}{cccc}
        
    \toprule
    \multirow{2}[2]{*}{$\lambda$} & \multicolumn{3}{c}{Predicates Classification} \\
\cmidrule{2-4}          & mR@20 & mR@50 & mR@100 \\
    \midrule
          0.00  &     25.24   &      31.95    &  34.44\\
        0.25  &      \textbf{30.97} &     35.91  &  37.88\\
        0.50  & 30.10 & \textbf{36.28} & \textbf{38.67}\\
        0.75  &     29.73  &    34.36   &  36.79\\
        1.00  &      29.20 &     33.47  &  35.69\\
    \bottomrule
    \end{tabular}%
      }
    
    \caption{\textbf{Ablation of $\lambda$}}
    \label{tab:bal on lambda}
    \end{subtable}

    \caption{\textbf{Ablation on MOTIFS}. We explore different numbers of layers and $\lambda$ with MOTIFS on the performance of Mean Recall@K.}

\end{table*}%

\subsection{Ablation Studies}
\label{sec:ablation}
\noindent\textbf{CR and MR}. We further explore the contributions of each term to our overall loss. In the SGG task, Recall@K and Mean Recall@K~\cite{c3} restrict mutually with each other. Recall@K represents how well the model performs for the predicate classes' head part. Thus, it reflects how well a model can imitate the previous model. On the contrary, Mean Recall@K~\cite{c3} evaluates the model's overall performance. Suppose we want to figure out the functionality of the knowledge consolidation term in the loss. In that case, it is reasonable to adopt Recall@K since two terms of knowledge reconstruction aim to prevent the model from forgetting previous knowledge. According to \cref{tab:up}, if we add CR and MR separately, mR@K will get constant improvement. However, only when CR and MR are used simultaneously will we get the highest mR@K and prevent R@K from dropping too much. Also, after comparing the second and third row of each model on R@K, it is obvious that MR is a more powerful constraint than CR.

\noindent\textbf{Layer Number}. The number of layers (i.e., stage) depends on how many top-K frequent predicates we pick up after clustering. We conduct experiments in \cref{tab:ab on ln} on different layers. We figured out that 2 is suitable for the VG dataset, mainly due to the small number of predicate classes and limited granularity variance.
HML training indeed needs more time to complete training during multi-stage training. Nevertheless, the increase ratio (125\%) of model performance is way more significant than the one (65\%) of training time. More time analysis will be shown in the supplementary material.
All experiments were carried out on one identical GPU.

\noindent\textbf{Hyperparameter $\lambda$}. In order to figure out the effect of $\lambda$ on the performance of the model, we set 5 values in ablation to $\lambda \in \{0, 0.25, 0.50, 0.75, 1.00\}$ in \cref{tab:bal on lambda}. $\lambda$ represents how much information will be passed down to the next stage. If $\lambda$ is too high, the new model will stick to the original classes without learning new ones. In contrast, low $\lambda$ can not guarantee effective information passing. Based on our experiment, $\lambda=0.5$ is suitable for the HML framework on VG.


\section{Conclusion}
We propose a general framework to enable SGG models to make fine-grained predictions. In addition to the objective long-tail effect in the dataset, we uncover mixed-granularity predicates caused by subjective human annotation. The similarity between the human hierarchical learning pattern and the SGG problem is obvious under this condition. Based on that, we designed the HML framework with two new constraints (i.e., CR and MR) for efficient training. We observe that the HML framework can improve performance compared to the traditional training fashion models and achieves new state-of-the-art.

\subsubsection{Acknowledgments}
This work was partly supported by the National Natural Science Foundation of China under Grant 41971284; the Fundamental Research Funds for the Central Universities under Grant 2042022kf1201; Wuhan University-Huawei Geoinformatics Innovation Laboratory. We sincerely thank our reviewers and ACs for providing insightful suggestions.

\clearpage
%
%
\bibliographystyle{splncs04}
\bibliography{egbib}
\end{document}